\pdfoutput=1

\documentclass[11pt]{article}

\usepackage[preprint]{coling}
\usepackage{times}
\usepackage{latexsym}

\usepackage{alphabeta}
\usepackage{svg}
\usepackage{textcomp}

\usepackage[T1]{fontenc}
\usepackage[utf8]{inputenc}

\usepackage{microtype}
\usepackage{makecell}
\usepackage{tabularx}
\usepackage{calc}
\usepackage{printlen}
\usepackage{enumerate}% http://ctan.org/pkg/enumerate

\usepackage{inconsolata}

\usepackage{graphicx}

\title{A State-of-the-Art Morphosyntactic Parser and Lemmatizer \\ for Ancient Greek}

\author{Giuseppe G. A Celano \\
Faculty of Mathematics and Computer Science\\
Institute of Computer Science \\
Leipzig University \\
\texttt{celano@informatik.uni-leipzig.de}}

\begin{document}
\maketitle
\begin{abstract}
This paper presents an experiment consisting in 
the comparison of six models
to identify a state-of-the-art morphosyntactic parser and 
lemmatizer for Ancient Greek capable of
annotating according 
to the Ancient Greek Dependency Treebank 
annotation scheme. A normalized
version of the major collections of
annotated texts was used to (i) train
the baseline model Dithrax with 
randomly initialized character embeddings 
and (ii) fine-tune 
Trankit and four recent models pretrained 
on Ancient Greek texts,
i.e., GreBERTa and PhilBERTa for morphosyntactic
annotation and GreTA and PhilTa for lemmatization.
A Bayesian analysis shows that Dithrax and Trankit
annotate morphology practically equivalently, 
while syntax is
best annotated by Trankit and lemmata by
GreTa. The results of the experiment 
suggest that token embeddings are 
not sufficient to achieve high UAS and LAS scores
unless they are coupled with 
a modeling strategy specifically designed
to capture syntactic relationships. The
dataset and best-performing models 
are made available online for reuse.

\end{abstract}

\section{Introduction}

In recent years, a few open-access 
annotated Ancient Greek (AG) corpora, such as
\textit{Opera Graeca Adnotata} (OGA) 
\citep{celano2024opera}
and the GLAUx corpus
\citep{keersmaekers2021glaux}, 
have been made available online. These
corpora allow for search of 
morphosyntax and lemmata across a wide variety of
AG texts, thus filling the gap of corpora such as
the \textit{Thesaurus Linguae Graecae}, whose
subscription-based access query engine 
is limited to word forms and lemmata.

Because of the token count 
in the order of millions,
the morphosyntactic annotation and lemmatization
of the above-mentioned open-access corpora
are feasible only if performed automatically. 
This raises a number of questions about 
which recent technology would be best suited for that purpose. 

OGA annotations rely 
on the COMBO parser \citep{rybak-wrblewska:2018:K18-2}, 
which, despite
being accurate,\footnote{
\href{https://git.informatik.uni-leipzig.de/celano/combo_for_ancient_greek}{https://git.informatik.uni-leipzig.de/celano/combo\_for\\\_ancient\_greek}} 
was built on Tensorflow 1
and is not actively maintained anymore. 
The GLAUx corpus employed RFTagger \citep{schmid-laws-2008-estimation}, 
Lemming \citep{muller-etal-2015-joint},
and the Stanford Graph-Based Dependency 
Parser 
\citep{dozat-etal-2017-stanfords} for annotation of, respectively, morphology,
lemmata, and syntax: the models
perform well \citep{keersmaekers2021glaux}, 
but have not been released,
and therefore cannot be reused.

For these reasons, the current paper 
presents a comparison of six models 
to identify and release
a state-of-the-art
morphosyntactic parser and lemmatizer that can 
annotate 
AG sentences according to the annotation scheme of
the Ancient Greek Dependency Treebank (AGDT).\footnote{
The best-performing models for morphosyntactic annotation
and lemmatization can be found at
\href{https://git.informatik.uni-leipzig.de/celano/morphosyntactic_parser_for_oga}{https://git.informatik.uni-leipzig.de/celano/morphosyntactic\\\_parser\_for\_oga} and
\href{https://git.informatik.uni-leipzig.de/celano/lemmatizer_for_oga}{https://git.informatik.uni-leipzig.de/\\celano/lemmatizer\_for\_oga},
respectively.
}
To promote future machine learning-based 
studies on AG,
the normalized version of the AG texts used for training
and now documented with their alleged composition dates 
for the first time is also released.\footnote{
\href{https://git.informatik.uni-leipzig.de/celano/morphosyntactic_parser_for_oga}{https://git.informatik.uni-leipzig.de/celano/morphosyn\\tactic\_parser\_for\_oga}
}

In Section \ref{relatedwork}, related work is
reviewed, while Section \ref{dataset}
describes the dataset used for training.
In Section \ref{experiment},
the experiment and the architectures of the different 
models compared are presented: the
results of their training 
are reported with a Bayesian statistical 
analysis in Section \ref{results} and
discussed in Section \ref{discussion}. Finally,
concluding remarks are contained in
Section \ref{conclusions}.

\section{Related work}
\label{relatedwork}

The explosion of machine learning in NLP 
has generated an ever-increasing 
number of resources, the reuse of which, 
however, is often not possible or 
straightforward 
due to the many different variables 
involved in each system.

Building on \citet{keersmaekers2021glaux},
\citet{keersmaekers2023creating}
documented
the parsing and lemmatization of
a large corpus consisting of literary 
and papyrological AG texts annotated according
to the AGDT annotation 
scheme. Interestingly,
they presented experiments
to increase LAS and UAS scores, 
in which original data were
transformed before training: 
for example, elliptical
nodes were deleted and 
the annotation style 
for coordination modified.
The reported results 
show some UAS and LAS increases 
in absolute terms. The models, however,
have not been released.

Most recent systems 
for morphosyntactic annotation
and lemmatization were trained on
the Universal Dependencies data, consisting
of two treebanks, the Perseus treebank
and the PROIEL treebank,\footnote{Recently, 
the PTNK treebank (about 39K tokens) 
has been added, but, as far as I know, 
it has not yet been used for machine learning experiments.} 
for a total of
about 416K tokens---notably,
the size of UD treebanks 
is less than half
of that of the data 
annotated with the
AGDT annotation scheme used in the
present study (see Section \ref{dataset}). 

The UD treebanks implement 
the UD annotation scheme differently,
and therefore creation of a single model
still represents a challenge:
\citet{kostkan2023odycy} 
provided a joint spaCy model 
for morphosyntactic annotation 
and lemmatization that seems to
 achieve good overall 
performance.\footnote{The
scores for the model \texttt{odyCy\_joint} 
on the UD Perseus treebank test set reported at 
\href{https://centre-for-humanities-computing.github.io/odyCy/performance.html}{https://centre-for-humanities-computing.github.io/odyCy/performance.html} 
are 
$95.39$ (POS tagging), $92.56$ (morphological features), 
$78.80$ (UAS), $73.09$ (LAS),
and $83.20$ (lemmatization). It is, however,
not clear whether the evaluation script used is 
the one of the CoNLL 2018 Shared Task (\href{https://universaldependencies.org/conll18/evaluation.html}{https://universaldependencies.org/conll18/evaluation.html}), 
which is commonly used in similar studies, including the present one. Since
this script does not allow for cycles 
and multiple roots, I suspect that 
the reported scores would be lower, if 
it had not been used.}

A number of studies reported on the creation of
token embeddings for AG 
by using the huge amount of texts
available online (\citealp{singh-etal-2021-pilot};
\citealp{yamshchikov-etal-2022-bert}).
Most recently, 
\citet{riemenschneider-frank-2023-exploring}
benchmarked a number of models 
for Ancient Greek and Latin. They show
that their
pretrained language model
GreBERTa achieves the (in absolute terms) 
highest performance 
scores for UPOS, XPOS, UAS, and LAS
when fine-tuned on the UD Perseus treebank
($95.83$, $91.09$, $88.20$, and $83.98$, respectively);
lemmatization is best performed by 
a T5 model they call GreTa, which
achieves an F1 score of $91.14$. 

\section{The dataset}
\label{dataset}

The dataset used for
training, validation, and testing 
consists of the following treebanks:
(i) the Ancient Greek Dependency Treebank,\footnote{\href{https://github.com/PerseusDL/treebank_data/releases}{https://github.com/PerseusDL/treebank\_data/releases}}
(ii) the Gorman Trees,\footnote{\href{https://github.com/vgorman1/Greek-Dependency-Trees}{https://github.com/vgorman1/Greek-Dependency-Trees}} and (iii) the
Pedalion Trees.\footnote{\href{https://github.com/perseids-publications/pedalion-trees}{https://github.com/perseids-publications/pedalion-trees}} 

All treebanks were natively
annotated using the AGDT annotation scheme, and
together they represent by far 
the largest morphosyntactically
annotated dataset for 
AG---and one of the largest
treebanks in absolute terms: the token count
of the texts before normalization is
$1,277,310$ and, after it, 
$1,260,863$.

As Table \ref{tdataset} shows, the final dataset comprises
a plethora of texts of different genre---including poetry, history,
and philosophy---and 
age, ranging from about 9th century BCE 
to 4th century CE (more details are provided in Appendix \ref{sec:appendixA}). Even though
the dataset is not balanced 
by genre and age, it is still representative
of most types of texts
written in the above-mentioned time span in Ancient Greece.

%\definecolor{babyblue}{rgb}{0.20, 0.8, 1}
%\definecolor{deepsaffron}{rgb}{1.0, 0.68, 0.18}
%\definecolor{myyellow}{rgb}{240, 225, 48}

\begin{table}[!ht]
  
  \centering

%\scalebox{0.8}{
\resizebox{\columnwidth}{!}{  

  \begin{tabular}{lccr}
    \hline
    \textbf{Author} & \textbf{Genre} & \textbf{Century} & \textbf{Tokens}\\
    \hline
    Hesiod, Homer & poem & $-9/8$ & $255,375$ \\ \hline
    \makecell[l]{Sappho,\\ Mimnermus,\\Semonides} & lyric     & $-7$ & $5,510$    \\\hline
    Homeric Hymns & hymns & $-7/6$ & $3,968$ \\ \hline
    Aesop    & fable & $-6$ & $5,221$ \\\hline
    \makecell[l]{Antiphon, Lysias,\\Isocrates} & oratory & $-5$   & $30,679$ \\\hline
    \makecell[l]{Aeschylus,\\Sophocles,\\Euripides} & tragedy & $-5$ & $108,386$ \\\hline
    \makecell[l]{Aristophanes,\\Cephisidorus\\Comicus} & comedy& $-5$ & $47,547$ \\\hline
    Aeneas Tacticus   & manual  & $-5$ & $7,207$ \\\hline
    \makecell[l]{Herodotus,\\Thucydides}     &  history & $-5$ & $65,494$ \\\hline
    Xenophon     & history & $-5/4$ & $142,635$ \\\hline
    \makecell[l]{Lysias, Isocrates,\\
    Demosthenes,\\Aeschines,\\ Andocides, Isaeus} & oratory & $-4$ & $153,088$ \\\hline
    \makecell[l]{Aristotle, Plato,\\Theophrastus}    & philosophy & $-4$ & $51,906$ \\\hline
    Menandrus & comedy & $-4$ & $8,069$ \\\hline

    Epicurus & philosophy & $-4/3$ & $1,523$ \\\hline
    
    Theocritus & lyric& $-3$ & $304$\\\hline
    Septuaginta & Bible & $-3$ & $19,235$ \\\hline
    Polybius & history & $-2$ & $105,693$ \\\hline
    Ezechiel & tragedy & $-2$ & $1,939$ \\\hline
    Batrachomyomachia & poem & $-1$ & $2,212$ \\\hline
    \makecell[l]{Diodorus of Sicily,\\ Dionysius\\of Halicarnassus} & history & $-1$ & $56,004$ \\ \hline
  Chion & epistolary & $+1$ & $5,577$ \\\hline

  Hero of Alexandria & science & $+1$ & 10,321\\\hline
  Josephus Flavius & history & $+1$ & $24,987$ \\\hline
  Chariton & romance & $+1/2$ & $6,265$ \\ \hline
    Plutarch & biography & $+1/2$ & $37,203$ \\\hline
  Phlegon of Tralles & paradox.& $+2$ & $5,642$ \\\hline
  Apollodorus & mythogr. & $+2$ & $1,265$ \\\hline
  Epictetus & philosophy & $+2$ & $7,204$ \\\hline
  Lucian  & novel & $+2$ & $11,054$ \\\hline
  Appianus & history & $+2$ & $25,665$ \\\hline
    Athenaeus& miscellany & $+2$ & $45,653$ \\\hline
  Longus & romance & $+2/3$ & $672$ \\\hline

  Sextus Empiricus & philosophy & $+3$ & $16,218$ \\\hline
  Paeanius & history & $+4$ & $6,184$ \\\hline
  Julian & oratory & $+4$ & $1,405$ \\\hline
 \end{tabular}
 }
  \caption{Statistics for the works contained in the dataset showing
  authors,
  genres, (alleged) 
  centuries of composition (indicated by Arabic numbers, with $+$ meaning CE and $-$ BCE), and token counts (before normalization). Full details in Appendix
  \ref{sec:appendixA}.}
  \label{tdataset}
\end{table}

\subsection{Normalization}

Since the final database consists of texts coming from different sources, which were annotated by many different scholars (sometimes adopting different conventions),
some automatic normalization of the original texts
was attempted to foster consistency and therefore
performance of machine learning algorithms.

Before training, all the relevant fields,
i.e., word form, lemma, POS tag, syntactic head and relation, needed some non-trivial format uniformation, especially to handle the case of
null or clearly erroneous 
values. Syntactic trees also had to be
modified if cycles were detected.

An often underestimated problem is that of character encoding for the apostrophe: all apostrophe-looking characters were
converted into the character MODIFIER LETTER APOSTROPHE (U+02BC), which affected about
50K characters.

While the vast majority of AG graphic words corresponds
to morphosyntactic tokens,\footnote{Crasis annotation, 
which is more elaborate to normalize,
was left untouched.} 
this is questionable 
for coordinate conjunctions such as, for example, οὐδὲ or εἴτε, which, in the final dataset, are tokenized
(therefore, οὐ δὲ and εἴ τε, respectively). Coordination
in the AGDT is not only annotated
at the level of syntactic tree but
also (redundantly) at that of 
syntactic label via use of the suffix
\texttt{\_CO}: to decrease 
the number of syntactic labels and
therefore supposedly improve algorithm performance, this and similar suffixes,
such as \texttt{\_AP}, were deleted.

Another related yet different
issue is represented by ellipsis, which poses a representational challenge. The AGDT annotation
scheme dates back to a pre-machine learning era, 
where elliptical nodes could be added
whenever they were necessary to 
build a syntactic tree. However, 
the complexity of the phenomenon 
and the absence of strict annotation rules 
on this matter led, 
over time, to the proliferation 
of various annotation styles: for example, 
sometimes the word form of an elliptical node
is specified, sometimes it is not. The
position of elliptical nodes within
a sentence is also problematic both
on a theoretical and representational level.

While \citet{keersmaekers2023creating}
proposed deletion of elliptical nodes,
\citeposs{celano2023neural} ellipsis modeling is
followed in the present study: elliptical
nodes were added at the end of a sentence
(whatever their alleged position was) and,
to avoid uncertainties about their word forms,
they were always encoded with placeholders such as
\texttt{[0]}, \texttt{[1]}, and so on, depending
on the number of them.\footnote{Since a model to
predict such elliptical nodes is provided at
\href{https://git.informatik.uni-leipzig.de/celano/ellipsis_Ancient_Greek}{https://git.informatik.uni-leipzig.de/celano/ellipsis\_Ancient\\\_Greek},
new texts can be made compliant to
this ellipsis annotation style.}

\begin{table*}[!ht]
  
\centering
\resizebox{\textwidth}{!}{
  
  \begin{tabular}{lccccccccccc}
    \hline
    \textbf{Model}  & \textbf{POS} & \textbf{XPOS} & \textbf{Feats} & \textbf{AllTags} & \textbf{UAS} & \textbf{LAS} & \textbf{Lemmas}\\
    \hline
    Dithrax   & $95.55\ (0.23)$ & $90.65\ (0.32)$ & $94.40\ (0.17)$ & $89.80\ (0.39)$ & $77.70\ (0.62)$ & $70.81\ (0.65)$ & $86.85\ (0.18)$ \\
    Trankit  &  $\mathbf{96.18\ (0.13)}$ & $\mathbf{91.55\ (0.21)}$ & $\mathbf{94.61\ (0.12)}$ & $\mathbf{91.21\ (0.22)}$ & $\mathbf{82.28\ (0.27)}$ & $\mathbf{76.67\ (0.34)}$ & N/A \\
    GreBERTa & $94.12\ (0.54)$  & $89.16\ (0.73)$ & $93.21\ (0.45)$ & $88.31\ (0.85)$ & $58.85\ (2.04)$ &$53.41\ (2.06)$ & N/A \\
    GreTa & N/A&N/A &N/A &N/A &N/A &N/A & $\mathbf{91.17\ (0.17)}$ \\
    PhilBERTa & $85.34\ (24.03)$ & $79.85\ (24.3)$ & $86.67\ (16.87)$ & $77.8\ (27.73)$ & $61.24\ (20.64)$ &$54.95\ (20.1)$ & N/A\\ 
    PhilTa & N/A&N/A &N/A &N/A &N/A &N/A &  $90.09\ (0.24)$  \\ \hline
    %GLAUx & N/A& $94.50$ &N/A &N/A & N/A& $79.43$ & $98.0$\\
    %\hline
    UD Perseus Trankit & $93.97$ & $87.25$ & $91.66$ &$86.88$
    &$83.48$ & $78.56$ & $88.52$ \\
    UD Perseus GreBERTa & $95.83$ & $91.09$ & N/A & N/A 
    & $88.20$ & $83.98$ & N/A\\
    UD Perseus GreTa + Chars & N/A&N/A &N/A &N/A &N/A &N/A & $91.14$  \\
    UD Perseus PhilBERTa & $95.60$ & $90.41$ & N/A & N/A
    & $86.99$ & $82.69$ & N/A \\
    UD Perseus PhilTa + Chars & N/A&N/A &N/A &N/A &N/A &N/A & $90.66$  \\
    \hline
  \end{tabular} 
  
  }
  \caption{
    Mean F1 scores + standard deviations in parentheses for 
    the test set results of the 5-fold cross-validation models (training on each split repeated twice with different random seeds). Best scores are in bold face. Results for parsers trained on UD Perseus data are
    shown only for loose comparison (see Section \ref{results}).
  }\label{scores}
\end{table*}

\section{Experiment}
\label{experiment}

A total of six model architectures 
were compared: four (i.e., three + baseline) for morphosyntactic
prediction and three (i.e., two + baseline) for lemma prediction.
More precisely, the baseline
model called \textbf{Dithrax}\footnote{The name derives from 
Dionysius Thrax, the author of the first extant
AG grammar.} is able to predict both 
morphosyntax and lemmata, while
the other five models 
can predict either one,
in that their modeling for character 
prediction for
lemmatization is kept
distinct from that for word prediction
for morphosyntax.

The performance of each model 
was evaluated using the evaluation script
used for the CoNLL 2018 Shared Task:
it outputs F1 scores for UPOS, XPOS, UFeats,
AllTags (i.e., UPOS+XPOS+UFeats), 
UAS (i.e., HEAD match), 
LAS (i.e., HEAD + DEPREL match), and Lemmas. 
Since the
AGDT tagsets are different, 
the above-mentioned metrics are conveniently renamed: 
POS, XPOS, Feats, AllTags, UAS, LAS, and Lemmas.

The original dataset was divided into
train, validation, and test sets (60\%, 20\%, 20\%).
Each model was trained 10 times, using 5-fold
cross-validation, with each training-validation split 
being used twice: as a result, 10 (i.e., $5\times2$) models
were trained for each model architecture (therefore, 10 final F1 scores
were calculated for each of the above-mentioned metrics). Since final models were not retrained on the entire dataset (train + validation sets) for time reasons, the scores presented (see Table \ref{scores}) are
the ones obtained on the test set---this allowed 
selection of the best model to use in production (see Table \ref{bestscores}). 

The training strategy is motivated by the fact that, 
while cross-validation
reduces variance by use of different splits of
the dataset, repetition of training 
on the same split allows for experimenting
with different random seeds.

\begin{table}[!]
\small
  \centering
  \begin{tabular}{lrrr}
    \hline
    \textbf{Model} & \textbf{APar} & \textbf{TPar} &\textbf{TTime}\\
    \hline
    Dithrax     & $58,906,077$ & $58,906,077$&${\approx}14.6$h\\
    Trankit & $283,463,421$ & $5,419,773$&${\approx}6.9$h\\ 
    GreBERTa & $127,860,506$ & $127,860,506$ & ${\approx}2.6$h\\
    GreTa     & $247,539,456$ &$247,539,456$&${\approx}11.4$h\\ 
    PhilBERTa & $137,076,506$ & $137,076,506$&${\approx}2.6$h\\
    PhilTa & $296,691,456$ & $296,691,456$&${\approx}12.3$h \\
    \hline
  \end{tabular}
  \caption{Model statistics for number of all parameters (APar),
  trainable parameters (TPar), and approximate 
  training time (TTime) calculated 
  on an Nvidia RTX4500 ADA 24GB GDDR6.}
  \label{ttimes}
\end{table}

\subsection{The statistical framework}

The results of the present experiment are
interpreted through the Bayesian analysis
proposed by \citet{JMLR:v18:16-305}.
More precisely, they propose 
a Bayesian correlated t-test to compare
cross-validation scores of two models  
on one dataset.

The proposed posterior distribution
coincides with the Student distribution used in the
frequentist t-test. This means that the
probabilities of the Bayesian correlated t-test 
coincide with the p-values of the frequentist
correlated t-test: what changes, however,
is the interpretation of such numerical values.

While the frequentist approach returns
the probability of data under the assumption that
the null hypothesis is true,
the Bayesian correlated t-test
computes the actual probabilities 
of the null and alternative hypotheses.

\citeposs{JMLR:v18:16-305} 
Bayesian correlated t-test 
provides three probability scores concerning
the comparison of the models $x$ and $y$ (see Section \ref{results}
for an example):

\begin{enumerate}[(i)]
\item $P(x = y)$: the probability of model $x$ being
practically equivalent to model $y$: this is the 
\textit{region of practical equivalence} (\textbf{rope})
corresponding to an arbitrary interval within which
two models are considered not to differ in practice. In the present study, this is $[{-}1, 1]$, i.e., the posterior probability of the
mean difference of F1 scores less than 1\% is considered
to mean practical equivalence.
\item $P(x \ll y)$: the probability that model $x$ is
practically worse than model $y$, i.e., 
the posterior probability of the 
mean difference 
of F1 scores being practically negative.
\item $P(x \gg y)$: the probability that model $x$ is
practically better than model $y$, i.e., 
the posterior probability of 
the mean difference 
of F1 scores being practically positive.
\end{enumerate}

The Bayesian approach offers 
a more straightforward 
statistical interpretation of data
and offers a solution for the well-known pitfalls 
of the frequentist framework, which include
the fact that null hypotheses are always
false in practice, and
therefore that enough data can obtain statistical 
significance even if the effect size is very small.

\begin{figure*}[ht]
\centering
\includegraphics[width=0.8\textwidth]{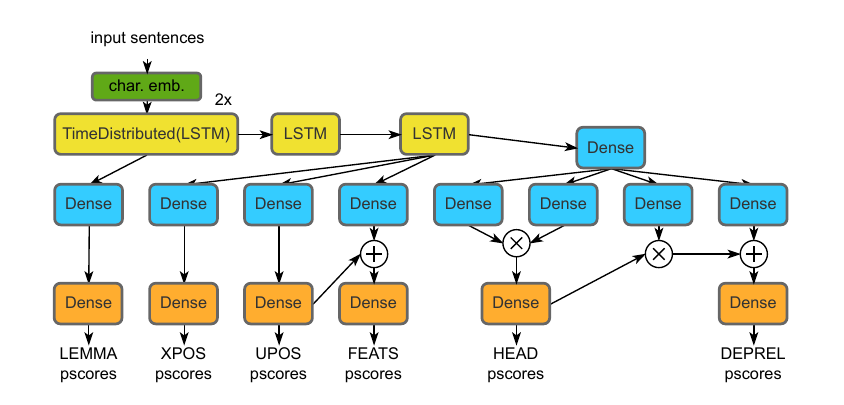}
\caption{Main layers of Dithrax, the baseline model architecture. Blue color stands for \texttt{tahn(linear(x))}, while orange for \texttt{softmax(linear(x))} (with $\times$ meaning dot product and $+$ concatenation).}
\label{dithrax}
\end{figure*}

\subsection{Dithrax: the baseline model}

As shown in Figure \ref{dithrax}, 
Dithrax is a multi-output 
LSTM model
vectorizing morphosyntactic tokens
into \textit{randomly initialized} 
character embeddings, which are
used for prediction of 
both lemmata and, after
further processing through LSTM layers, 
morphosyntactic labels. 

The model is inspired to the COMBO 
parser \citep{rybak-wrblewska:2018:K18-2},
which performed as one of the most accurate parsers
at the CoNLL 2018 Shared Task \citep{zeman-etal-2018-conll}. 

More precisely,
Dithrax proposes a
similar modeling strategy 
for HEAD and DEPREL 
labels based on
adjacency matrices
resulting from dot products of
two 2-rank tensors representing,
respectively,
heads and dependents of
the same sentence,
with each matrix row corresponding to
the vector representation of a 
token. 

\subsection{Trankit}

Trankit is a state-of-the-art 
transformer-based toolkit
for morphosyntactic analysis
and lemmatization. It is designed for UD data,
and is also able to process
raw documents, in that it
comprises a tokenizer
and sentence splitter. Key
features of Trankit are:
\begin{enumerate}[(i)]
    \item use of the multilingual pretrained 
transformer XLM-Roberta,
whose output is fine-tuned
on new data.
\item adapters: feed-forward networks
for each major component of Trankit (six in total),
whose weights---together with the specific ones for
final predictions---are updated only, while the
pretrained transformer ones remain fixed.
These make Trankit memory- and time-efficient.
\item syntax is modeled via
\citeposs{dozat2017deep} Biaffine
Attention.
\end{enumerate}

For the purpose of the present experiment,
we trained Trankit's joint model for part-of-speech tagging
, morphological feature tagging, 
and dependency parsing 
(i.e., POS, XPOS, Feats, AllTags, UAS, and LAS scores); the
lemmatizer could not be trained because of an internal
code error.

\subsection{Pretrained models: Gre(BERTa|Ta) and Phil(BERTa|Ta)}

The pretrained models GreBERTa and GreTa
(for AG) and 
PhilBERTa and PhilTa (for AG and Latin) were fine-tuned
for comparison,\footnote{
I use the names GreBERTa, PhilBERTa, GreTa, and PhilTa 
to also name the models resulting from the fine-tuning of the
homonymous pretrained embeddings: context is sufficient to
clarify what these names exactly refer to.} in that they
have recently been argued to perform
better than previous pretrained AG models
(see
\citealp{riemenschneider-frank-2023-exploring}
and Table \ref{scores}).

GreBERTa and GreTa were fine-tuned on
the Greek data of the Open Greek  and Latin Project,
the CLARIN corpus of Greek Medieval Texts,
the Patrologia Graeca, and the Internet Archive (in total, 
about 185.1M tokens).

PhilBERTa and PhilTa were fine-tuned on not only
AG but also Latin and English data.
The latter come from the the Corpus Corporum
project (167.5M tokens) and a collection of English texts
from different sources (212.8M tokens), whose topics
are similar to the ones found in AG and
Latin sources (for example, English translations
of AG and Latin texts), for a total
of 565.4M tokens.

GreBERTa and PhilBERTa are encoder-only
transformers 
providing token embeddings
for prediction of word-related
labels (i.e., UPOS, XPOS, UFeats, AllTags,
UAS, and LAS). Since not the original scripts 
but only the pretrained models 
are made available online 
(see also Section \ref{limitations}), it was not
possible to test the former with
the AGDT dataset (see Section \ref{discussion}):
in the present experiment, therefore, the pretrained
token embeddings were just used as inputs to
dense layers outputting the final probability
scores for each token.
GreTa and PhilTa are encoder-decoder 
transformers for
character prediction, and are therefore
 used for lemmatization.\footnote{I am grateful to
Frederick Riemenschneider, who provided me with
a script for lemma prediction similar to
the one used for his paper.}

\begin{table}[h]

\resizebox{\columnwidth}{!}{

  \begin{tabular}{lccccccccccc}
    \hline
    \textbf{Model}  & \textbf{POS} & \textbf{XPOS} & \textbf{Feats} & \textbf{AllTags} & \textbf{UAS} & \textbf{LAS} & \textbf{Lemmas}\\
    \hline
    Trankit   & $96.41$ & $91.90$ & $94.77$ & $91.56$ & $82.60$ & $77.10$ & N/A \\
    GreTa & N/A & N/A & N/A & N/A & N/A & N/A &  $91.41$ \\\hline
  \end{tabular}   
}
  \caption{
    Scores of the best-performing cross-validation runs evaluated on the test set.
  }\label{bestscores}
\end{table}

\section{Results}
\label{results}

Table \ref{scores} shows the mean F1 scores
and related standard deviations\footnote{SDs have been calculated using \texttt{numpy.std with ddof=1}.} for the models trained with 5-fold cross-validation, with each split being used
twice with different random seeds (in total, 10 models for each architecture). The mean scores are based on the F1 scores 
returned by 
the evaluation script of the CoNLL 2018 Shared Task
applied to the results outputted by each model when
tested on the test set.\footnote{New models were not trained on the train plus validation set for time reasons: therefore, performance of each model on the test set enabled selection of the best model to use in production.} The models created
by the runs with the best scores (see Table \ref{bestscores}) are made
available online.\footnote{\href{https://git.informatik.uni-leipzig.de/celano/morphosyntactic_parser_for_oga}{https://git.informatik.uni-leipzig.de/celano/morphosyn\\tactic\_parser\_for\_oga};
\href{https://git.informatik.uni-leipzig.de/celano/lemmatizer_for_oga}{https://git.informatik.uni-leipzig.de/\\celano/lemmatizer\_for\_oga}}

\begin{figure}[!ht]

\includegraphics[width=\columnwidth]{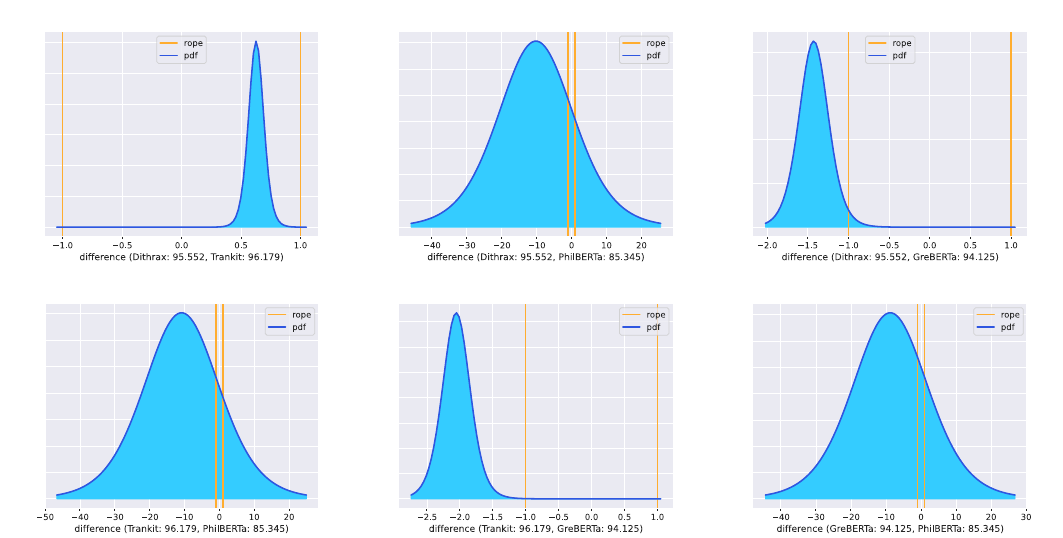}

\caption{Posteriors of the Bayesian correlated t-test for all
model pairs with reference to POS scores.}
\label{pos}
\end{figure}

\begin{table}[!ht]
\resizebox{\columnwidth}{!}{
  \begin{tabular}{lccc}
   
    \hline
    \textbf{Model pair}  & \textbf{Left} & \textbf{Rope} & \textbf{Right} \\
    \hline
    Dithrax-Trankit & ${\approx}0.00$ & $\mathbf{{\approx}1.00}$ & ${\approx}0.00$\\
    Dithrax-PhilBERTa & ${\approx}0.79$ & ${\approx}0.04$ & ${\approx}0.17$ \\
    Dithrax-GreBERTa & ${\approx}0.98$& ${\approx}0.02$ & ${\approx}0.00$ \\
    Trankit-PhilBERTa & $\mathbf{{\approx}0.80}$ & ${\approx}0.04$ & ${\approx}0.16$ \\
    Trankit-GreBERTa & $\mathbf{{\approx}1.00}$ & ${\approx}0.00$ & ${\approx}0.00$ \\

    GreBERTa-PhilBERTa & ${\approx}0.75$ & ${\approx}0.05$ & ${\approx}0.20$\\
    \hline
  \end{tabular}
  }
  \caption{
    Integrals on the intervals $(-\infty\, -1)$, 
    $[-1, 1]$, and $(1, +\infty)$ for plots
    in Figure \ref{pos}.
  }\label{posvalues}
\end{table}

Table \ref{scores} also displays the results for 
the models trained on
UD Perseus data, i.e., a small subset of
the dataset used for the present study, which were evaluated using the same
CoNLL 2018 Shared Task script.\footnote{Results for Trankit come from \href{https://trankit.readthedocs.io/en/latest/performance.html}{https://trankit.readthedocs\\.io/en/latest/performance.html} (\texttt{Ancient\_Greek-Perseus} treebank).}
Even if the UD annotation scheme and the 
AGDT one are similar, there are differences that are 
likely to impact parsing results. For example,
\citet{keersmaekers2021glaux} argues
that UD annotation style of coordination 
allows achieving higher scores for UAS and LAS.
Moreover, UD data, differently
from the AGDT data used for the present study,
do not contain elliptical nodes.
This means that comparison of F1 scores between
UD models and the ones of the present study can only
be loose, especially with reference to UAS and LAS.

\begin{figure}[!ht]

\includegraphics[width=\columnwidth]{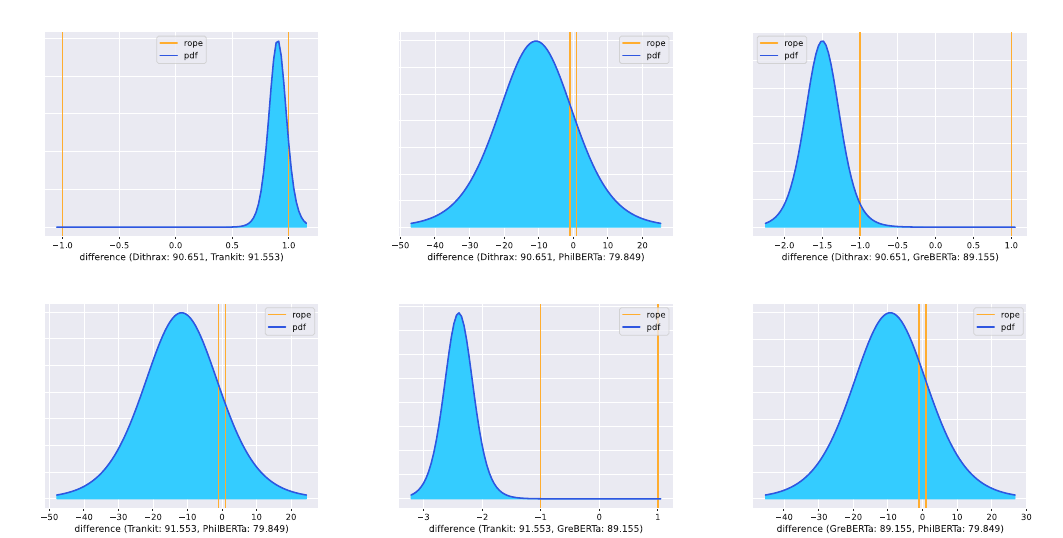}
\caption{Posteriors of the Bayesian correlated t-test for all
model pairs with reference to XPOS scores.}
\label{xpos}
\end{figure}

\begin{table}[!ht]
\resizebox{\columnwidth}{!}{
  \begin{tabular}{lccc}
   
    \hline
    \textbf{Model pair}  & \textbf{Left} & \textbf{Rope} & \textbf{Right} \\
    \hline
    Dithrax-Trankit & ${\approx}0.00$ & $\mathbf{{\approx}0.88}$ & ${\approx}0.12$\\

    Dithrax-PhilBERTa & ${\approx}0.80$ & ${\approx}0.04$ & ${\approx}0.16$ \\
    Dithrax-GreBERTa & ${\approx}0.97$ & ${\approx}0.03$ & ${\approx}0.00$ \\
    Trankit-PhilBERTa & $\mathbf{{\approx}0.82}$ & ${\approx}0.04$ & ${\approx}0.14$ \\
    Trankit-GreBERTa & $\mathbf{{\approx}1,00}$ & ${\approx}0.00$ & ${\approx}0.00$ \\

    GreBERTa-PhilBERTa & ${\approx}0.76$ & ${\approx}0.05$ & ${\approx}0.19$\\
    \hline
  \end{tabular}
  }
  \caption{
    Integrals on the intervals $(-\infty, -1)$, 
    $[-1, 1]$, and $(1, +\infty)$ for plots
    in Figure \ref{xpos}.
  }\label{xposvalues}
\end{table}

The mean scores for PhilBERTa shown in Table \ref{scores} are the lowest ones and their
related standard deviations are remarkably high (${>}20$) because
the model performed very poorly in one run. However, even if
that run were not considered,
the mean scores would still
be lower and the standard deviations rather high in comparison to the values
of the other models:
POS: $92.87\ (3.63)$; XPOS: $87.42\ (4.37)$;
Feats: $91.92\ (3.11)$; AllTags: $86.44\ (4.94)$; UAS:
$67.44\ (6.9)$; LAS: $60.94\ (7.11)$.

\begin{figure}[!ht]
\includegraphics[width=\columnwidth]{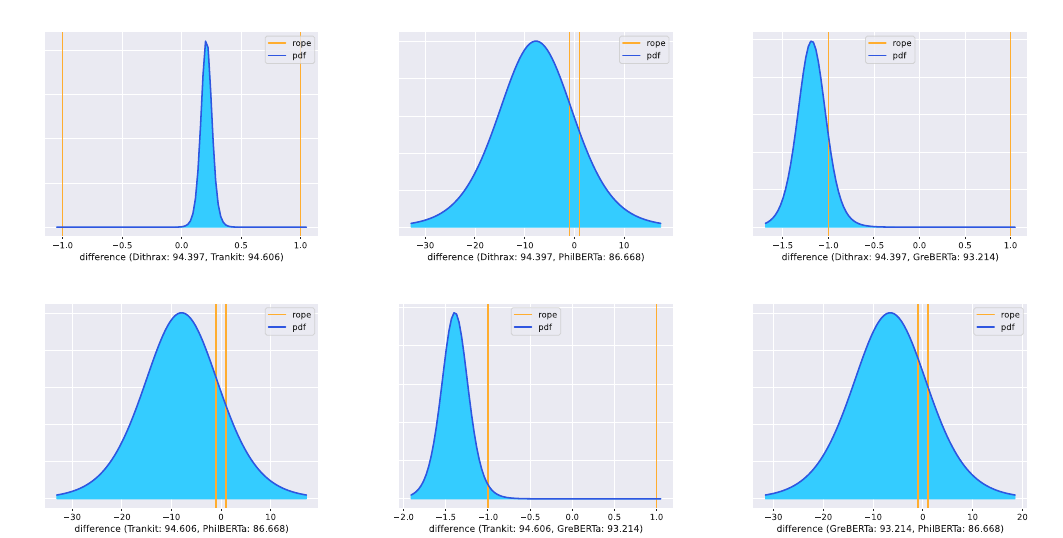}
\caption{Posteriors of the Bayesian correlated t-test for all
model pairs with reference to Feats scores.}
\label{feats}
\end{figure}

\begin{table}[!ht]
\resizebox{\columnwidth}{!}{
  \begin{tabular}{lccc}
   
    \hline
    \textbf{Model pair}  & \textbf{Left} & \textbf{Rope} & \textbf{Right} \\
    \hline
    Dithrax-Trankit & ${\approx}0.00$ & $\mathbf{{\approx}1.00}$ & 
    ${\approx}0.00$ \\
    Dithrax-PhilBERTa & ${\approx}0.80$ & 
    ${\approx}0.06$ & 
    ${\approx}0.14$ \\
    Dithrax-GreBERTa & ${\approx}0.86$ & ${\approx}0.14$ & ${\approx}0.00$ \\
    Trankit-PhilBERTa & $\mathbf{{\approx}0.80}$ & ${\approx}0.06$ & ${\approx}0.14$ \\
    Trankit-GreBERTa & $\mathbf{{\approx}0.98}$ & ${\approx}0.02$ & ${\approx}0.00$ \\

    GreBERTa-PhilBERTa & ${\approx}0.75$ & ${\approx}0.07$ & ${\approx}0.18$ \\
    \hline
  \end{tabular}
  }
  \caption{
    Integrals on the intervals $(-\infty, -1)$, 
    $[-1, 1]$, and $(1, +\infty)$ for plots
    in Figure \ref{feats}.
  }\label{featsvalues}
\end{table}

\begin{figure}[!ht]
\includegraphics[width=\columnwidth]{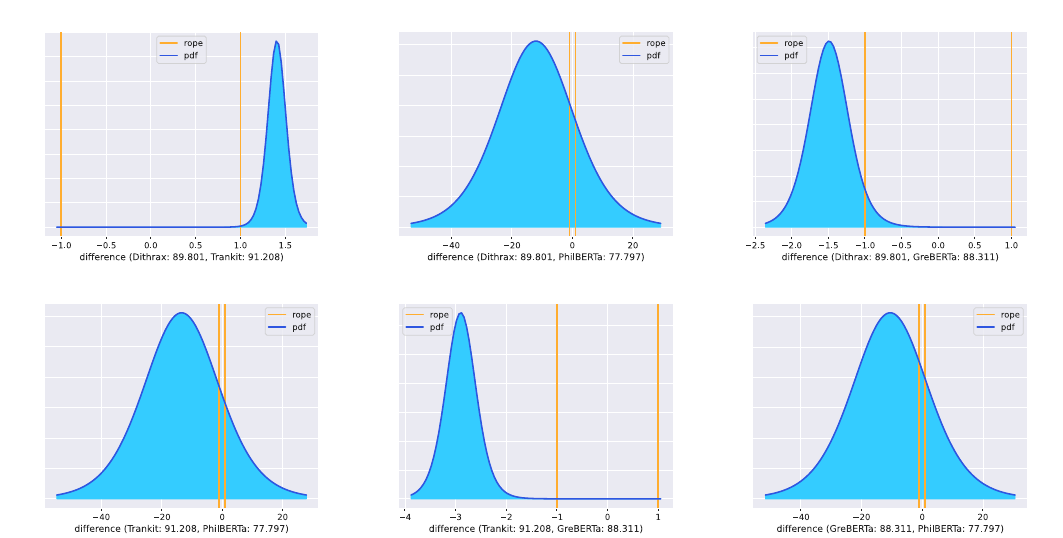}

\caption{Posteriors of the Bayesian correlated t-test for all
model pairs with reference to AllTags scores.}
\label{alltags}
\end{figure}

\begin{figure}[!ht]
\includegraphics[width=\columnwidth]{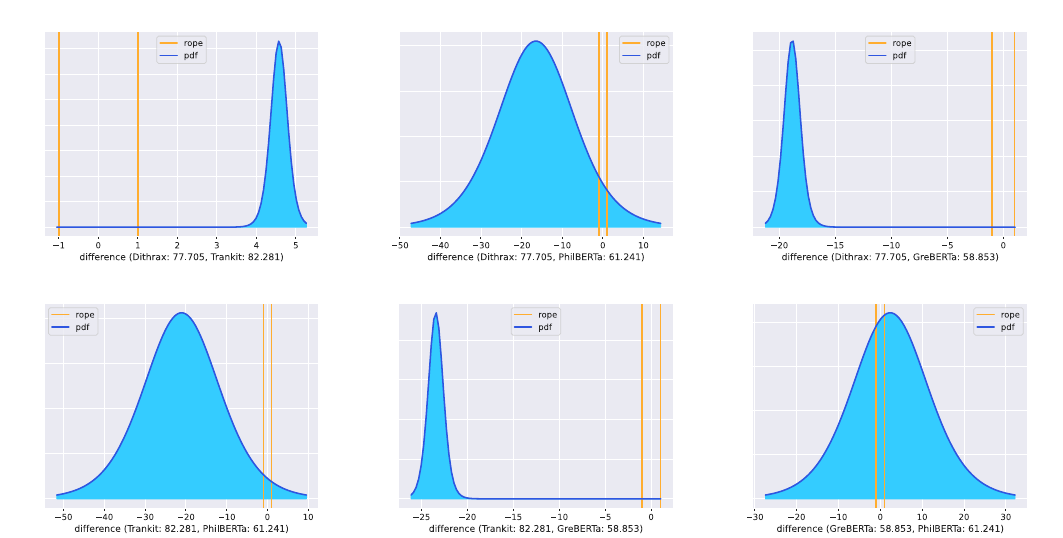}

\caption{Posteriors of the Bayesian correlated t-test for all
model pairs with reference to UAS scores.}
\label{uas}
\end{figure}

\begin{figure}[!ht]
\includegraphics[width=\columnwidth]{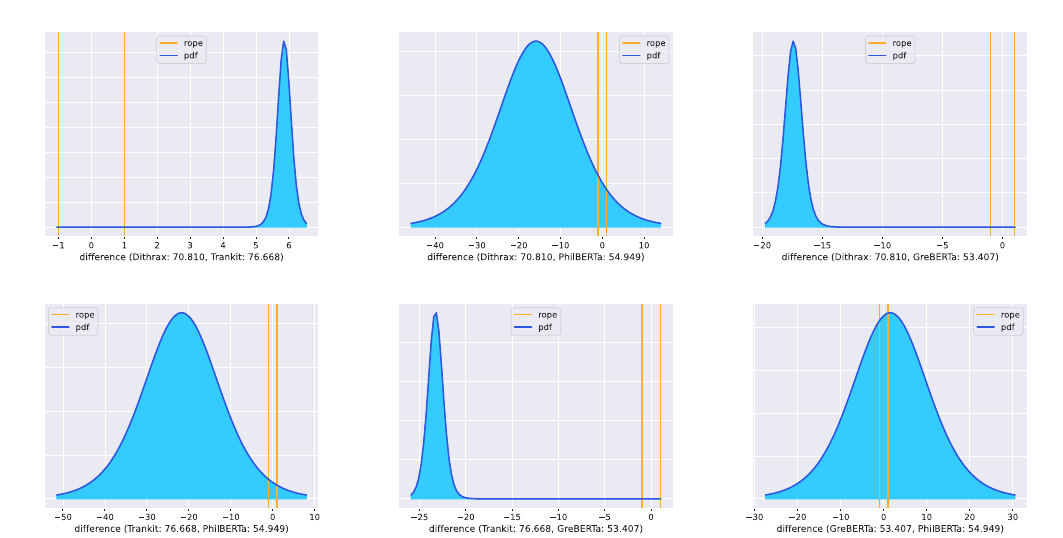}

\caption{Posteriors of the Bayesian correlated t-test for all
model pairs with reference to LAS scores.}
\label{las}
\end{figure}

\begin{figure}[!ht]
\centering
\includegraphics[width=\columnwidth]{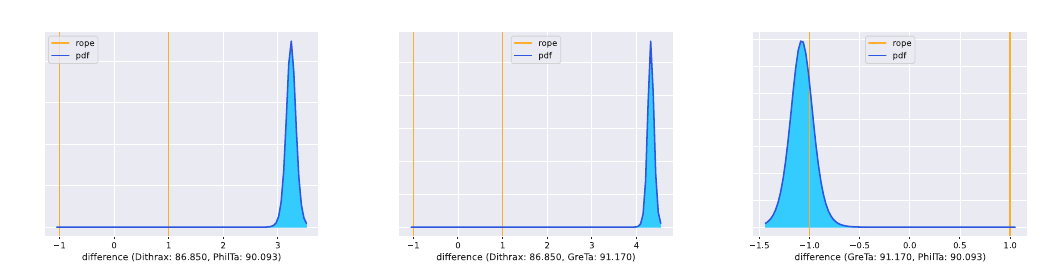}

\caption{Posteriors of the Bayesian correlated t-test for all
model pairs with reference to Lemmas scores.}
\label{lemmas}
\end{figure}

Figures \ref{pos}, \ref{xpos},
\ref{feats}, \ref{alltags},
\ref{uas}, \ref{las}, and \ref{lemmas} show
the posterior distributions of the mean differences of F1 scores between 
all models pairwise returned by
\citeposs{JMLR:v18:16-305} Bayesian correlated t-test.\footnote{The Python package documented at \href{https://github.com/janezd/baycomp}{https://github.com/\\janezd/baycomp} was used for the plots and calculations.}

\begin{table}[!ht]
\resizebox{\columnwidth}{!}{
  \begin{tabular}{lccc}
   
    \hline
    \textbf{Model pair}  & \textbf{Left} & \textbf{Rope} & \textbf{Right} \\
    \hline
    Dithrax-Trankit & ${\approx}0.00$ & ${\approx}0.00$ & $\mathbf{{\approx}1.00}$\\
    Dithrax-PhilBERTa & ${\approx}0.80$ & ${\approx}0.04$ & ${\approx}0.17$ \\
    Dithrax-GreBERTa & ${\approx}0.95$ & ${\approx}0.05$ & ${\approx}0.00$ \\
    Trankit-PhilBERTa & $\mathbf{{\approx}0.82}$ & ${\approx}0.03$ & ${\approx}0.14$ \\
    Trankit-GreBERTa & $\mathbf{{\approx}1.00}$ & ${\approx}0.00$ & ${\approx}0.00$ \\

    GreBERTa-PhilBERTa & ${\approx}0.76$ & ${\approx}0.04$ & ${\approx}0.19$\\
    \hline
  \end{tabular}
  }
  \caption{
    Integrals on the intervals $(-\infty, -1)$, 
    $[-1, 1]$, and $(1, +\infty)$ for plots
    in Figure \ref{alltags}.
  }\label{alltagsvalues}
\end{table}

In each of the above-mentioned figures except Figure \ref{lemmas}, 
the top-left, top-middle, top-right,
bottom-left, bottom-middle, bottom-right plots show, respectively, the posteriors for the pairs Dithrax-Trankit, Dithrax-PhilBERTa, Dithrax-GreBERTa, Trankit-PhilBERTa,
Trankit-GreBERTa, and
GreBERTa-PhilBERTa. In Figure \ref{lemmas}, which
visualizes Lemmas scores, the left, middle, and right plots
represent the posteriors for Dithrax-PhilBERTa, Dithrax-GreBERTa,
and GreBERTa-PhilBERTa, respectively---Trankit
could not be trained for lemmatization 
because of an internal error code. Each above-mentioned
Figure is coupled with a table (i.e., Tables \ref{posvalues}, \ref{xposvalues},
\ref{featsvalues}, \ref{alltagsvalues},
\ref{uasvalues}, \ref{lasvalues}, and \ref{lemmasvalues}), 
which reports the values of the areas under the curves. 

\begin{table}[!ht]
\resizebox{\columnwidth}{!}{
  \begin{tabular}{lccc}
    \hline
    \textbf{Model pair}  & \textbf{Left} & \textbf{Rope} & \textbf{Right} \\
    \hline
    Dithrax-Trankit & ${\approx}0.00$ & ${\approx}0.00$ & $\mathbf{{\approx}1.00}$\\
    Dithrax-PhilBERTa & ${\approx}0.93$ & ${\approx}0.02$ & ${\approx}0.05$ \\
    Dithrax-GreBERTa & ${\approx}1.00$& ${\approx}0.00$ & ${\approx}0.00$ \\
    Trankit-PhilBERTa & 
    $\mathbf{{\approx}0.97}$ & ${\approx}0.01$ & ${\approx}0.02$ \\
    Trankit-GreBERTa & 
    $\mathbf{{\approx}1.00}$ & ${\approx}0.00$ & ${\approx}0.00$ \\

    GreBERTa-PhilBERTa & ${\approx}0.36$ & ${\approx}0.08$ & ${\approx}0.56$\\
    \hline
  \end{tabular}
  }
  \caption{
    Integrals on the intervals $(-\infty, -1)$, 
    $[-1, 1]$, and $(1, +\infty)$ for plots
    in Figure \ref{uas}.
  }\label{uasvalues}
\end{table}

Each single plot gives information
about the probabilities that the mean differences
of F1 scores between two models are practically negative, practically equivalent, and practically positive. For example,
the bottom-left plot in Figure \ref{feats}
and the corresponding Table \ref{featsvalues}
show:

\begin{itemize}
  \item the posterior probability that the mean difference of F1 scores between PhilBERTa and Trankit is practically negative, i.e., the integral of the posterior on the interval $(-\infty, -1)$, equaling to ${\approx}0.80$. This is the probability that
  Trankit is practically \textbf{better} than PhilBERTa.
  \item the posterior probability that the mean difference of F1 scores between PhilBERTa and Trankit is practically equivalent, i.e., the integral of the posterior on the \textit{rope} interval $[-1, 1]$, equaling to ${\approx}0.06$. This is the probability that PhilBERTa and Trankit are practically \textbf{equivalent}.
\item the posterior probability that the mean difference of F1 scores between PhilBERTa and Trankit is practically positive, i.e., the integral of the posterior on the interval 
$(1, +\infty)$, equaling to ${\approx}0.14$. This is the probability that PhilBERTa is practically \textbf{better} than Trankit.
\end{itemize}

\begin{table}[!ht]
\resizebox{\columnwidth}{!}{
  \begin{tabular}{lccc}
   
    \hline
    \textbf{Model pair}  & \textbf{Left} & \textbf{Rope} & \textbf{Right} \\
    \hline
    Dithrax-Trankit & ${\approx}0.00$ & ${\approx}0.00$ & $\mathbf{{\approx}1.00}$\\
    Dithrax-PhilBERTa & ${\approx}0.93$ & ${\approx}0.02$ & ${\approx}0.05$ \\
    Dithrax-GreBERTa & ${\approx}1.00$& ${\approx}0.00$ & ${\approx}0.00$ \\
    Trankit-PhilBERTa & $\mathbf{{\approx}0.98}$ & ${\approx}0.01$ & ${\approx}0.02$ \\
    Trankit-GreBERTa & $\mathbf{{\approx}1.00}$ & ${\approx}0.00$ & ${\approx}0.00$ \\
    GreBERTa-PhilBERTa & ${\approx}0.39$ & ${\approx}0.09$ & ${\approx}0.52$\\
    \hline
  \end{tabular}
  }
  \caption{
    Integrals on the intervals $(-\infty, -1)$, 
    $[-1, 1]$, and $(1, +\infty)$ for plots
    in Figure \ref{las}.
  }\label{lasvalues}
\end{table}

\begin{table}[!ht]
\centering
  \begin{tabular}{lccc}
    \hline
    \textbf{Model pair}  & \textbf{Left} & \textbf{Rope} & \textbf{Right} \\
    \hline
    Dithrax-PhilTa & ${\approx}0.00$ & ${\approx}0.00$ & ${\approx}1.00$ \\
    Dithrax-GreTa & ${\approx}0.00$& ${\approx}0.00$ & $\mathbf{{\approx}1.00}$ \\
    GreTa-PhilTa & $\mathbf{{\approx}0.75}$ & ${\approx}0.25$ & ${\approx}0.00$\\
    \hline
  \end{tabular}
  \caption{
    Integrals on the intervals $(-\infty, -1)$, 
    $[-1, 1]$, and $(1, +\infty)$ for plots
    in Figure \ref{lemmas}.
  }\label{lemmasvalues}
\end{table}

\section{Discussion}
\label{discussion}

Table \ref{scores} seems to suggests that Trankit is
the best model in each morphosyntactic task. 
This is
only \textit{partly} confirmed by the Bayesian statistical analysis.

Even if Trankit's results for POS, XPOS, and Feats
are the highest in absolute terms, 
its performance can be considered to be
practically equivalent to that of
the baseline model Dithrax with reference to these metrics. 

Indeed,
the corresponding (top-left) plots in Figure \ref{pos},
\ref{xpos}, and \ref{feats} show that the area of
the t-curve within the rope is ${\approx}1$ for POS and Feats and ${\approx}0.88$ for XPOS. On the other hand, the models PhilBERTa and GreBERTa perform practically
worse than both Dithrax and Trankit (see top-middle,
top-right, bottom-left, and bottom-middle plots) with respect to these same metrics:
there is at least a ${\approx}0.79$ (see Dithrax-PhilBERTa in Table \ref{posvalues}) or higher probability that Dithrax or
Trankit perform practically better.

This is an interesting result because, differently
from Trankit, PhilBERTa, and GreBERTa, Dithrax does 
not rely on pretrained (but randomly initialized) 
(character) embeddings and its architecture has
a lower overall number of 
parameters (see Table \ref{ttimes}):\footnote{Trankit has
fewer trainable parameters than Dithrax, but unlike Dithrax, 
it only predicts morphosyntax.
} this suggests
that classification tasks such as POS, XPOS, and
Feats can be successfully addressed without 
use of more expensive model architectures.

The AllTags F1 score is a metric for 
POS+XPOS+Feats. Trankit turns out to
perform practically better than any other model (Figure \ref{alltags}),
including Dithrax, even if the top-left
plot shows that most area under the curve is
in $(1, +\infty)$, i.e.,
close to the rope interval endpoint: this reflects the fact that the mean difference between the F1 values of Trankit and
Dithrax is ${\approx}1.41$.

Syntactic prediction is notoriously more
complex, and this is shown in the lower results
reported in Table \ref{scores} for UAS and LAS.
Trankit's performance is clearly superior
than that of any other model, even if its scores
are much lower than the POS and XPOS ones.

Syntactic analysis is a much more challenging task
because HEAD and DEPREL values heavily depend
on contextual information. Even if
a pretrained transformer such as
GreBERTa or PhilBERTa outputs
context-aware token embeddings,
they turn out to poorly predict syntax
without a further modeling strategy.

In the GreBERTa and PhilBERTa models,
the pretrained token embeddings were
used as an input to dense layers
outputting probabilities for
morphology and syntax in a 
multi-output model: however,
while results for morphology
are comparable to those of the other models,
those for syntax clearly are not (see also Section \ref{limitations}):
as Table \ref{scores} shows,
UAS and LAS scores for GreBERTa and PhilBERTa 
are remarkably lower, and there is
a ${\approx}0.93$ or higher probability that
Dithrax or Trankit perform practically better than
them (Tables \ref{uasvalues} and \ref{lasvalues}). 

This can be explained by the fact that,
contrary to GreBERTa and PhilBERTa,
Dithrax and Trankit employ a modeling strategy
on the top of embeddings:
Dithrax models sentence syntax through
adjacency matrices \citep{rybak-wrblewska:2018:K18-2}, while
Trankit implements \citeposs{dozat2017deep}
biaffine attention mechanism, both of which
aim to capture the complex 
relationship
between heads and dependents 
within a sentence.\footnote{To filter syntactic cycles, the Chu-Liu-Edmonds algorithm
is applied to each parser's output.}

Lemmatization is performed best by
GreTa. While Dithrax
simply employes LSTM layers over
character embeddings, GreTa and PhilTa
are Seq2seq models: Table \ref{lemmasvalues}
shows that, while the Seq2seq models
perform practically better than 
Dithrax (${\approx}1.00$), there is a ${\approx}0.75$ probability
that GreTa performs practically better than
PhilTa and a ${\approx}0.25$ probability
that their performance is practically
equivalent.

If we compare Trankit's results on the AGDT dataset
with those on the UD dataset (see Table \ref{scores}),
scores for POS, XPOS, Feats, and AllTags
are considerably higher in absolute terms on the AGDT dataset, with
differences of ${\approx}2.21$, ${\approx}4.3$,
${\approx}2.95$, and ${\approx}4.33$, respectively;
UAS and LAS scores, however, are higher on the
UD dataset, with differences of ${\approx}1.2$ and
${\approx}1.89$, respectively. Interestingly,
UAS and LAS scores do not seem 
to be impacted by the much larger size 
of the AGDT dataset, even if
the model trained on AGDT data can be expected to
generalize much better than that on the UD data due
to the much larger variety of its texts.

\section{Conclusions}
\label{conclusions}

A comparison of six model architectures
(Dithrax, Trankit, PhilBERTa, GreBERTa, PhilTa, and GreTa)
was
documented to select state-of-the-art
models for annotation of morphosyntax and
lemmata according to the AGDT annotation scheme. A Bayesian
statistical analysis was adopted to
interpret cross-validation scores, which
suggests that Trankit annotates syntax 
better than the other models do,
while GreTa's performance for lemmatization is 
the best. The study shows that 
the baseline model Dithrax can also
achieve state-of the-art performance 
for morphological annotation, although
it employs a lower overall number of 
parameters and randomly 
initialized character 
embeddings. 

A noteworthy 
finding of the study is
that, even if pretrained embeddings, such as
GreBERTa and PhilBERTa, 
rely on complex model architectures
vectorizing tokens on the basis
of their linguistic context, which
were calculated on very large collections
of AG texts, they do not perform well for
syntactic prediction (i.e., UAS and LAS scores),
unless a further modeling strategy aimed at
capturing syntax information within a sentence
is put in place, such as adjacency matrices
or biaffine attention.

\section{Limitations}
\label{limitations}

The study aimed to document
a state-of-the-art model for
morphosyntactic analysis and lemmatization
of Ancient Greek. The dataset
used for training contains manual annotations
that were performed over many years
by different (single) annotators 
(some were students, others
scholars). Therefore, as is often the
case with manual annotations, annotation 
consistency within the dataset is not guaranteed 
because of either annotation errors
or different annotation styles, the
first annotation guidelines\footnote{\href{https://github.com/PerseusDL/treebank_data/blob/master/v1/greek/docs/guidelines.pdf}{https://github.com/PerseusDL/treebank\_data/blob/master/\\v1/greek/docs/guidelines.pdf}; newer annotated texts (should) 
follow the much more specific annotation guidelines at
\href{https://github.com/PerseusDL/treebank_data/blob/master/AGDT2/guidelines/}{https://github.com/PerseusDL/treebank\_data/blob/master/\\AGDT2/guidelines/}} not being
specific on a number of morphosyntactic
phenomena---it should also be noted that
morphosyntactic annotation of Ancient Greek
literary texts 
is arguably much more complex than that 
of modern texts.

For this reason, the present study set aside 
the issue of how annotation quality/consistency 
affects parsing results. Similarly, no experiment
was conducted with reference to corpus composition,
under the assumption that model architectures 
are powerful enough
to capture distinctions between texts of different
genre and/or composition dates.

Reuse of 
models and model architectures for comparison was often limited:
they are either not released or the provided code
is partial. The latter case is that of PhilBERTa
and GreBERTa: they
achieved state-of-the-art UAS and LAS scores on the UD Perseus treebank, 
but the original scripts have not (yet) been released,\footnote{\href{https://github.com/Heidelberg-NLP/ancient-language-models/tree/main}{https://github.com/Heidelberg-NLP/ancient-language-models/tree/main}.} and therefore their original model architectures could not be used in the present study.

\section*{Acknowledgments}

This work is supported by the German Research Foundation (DFG project number
408121292).
\bibliography{custom}

\appendix

\section{Texts}
\label{sec:appendixA}

The following tables provide details of 
the texts used for train, validation, 
and test sets (see also Table \ref{dataset}
for a more concise presentation). 
The authors, title, and dates of each work 
were retrieved mostly from the file
\href{https://github.com/OperaGraecaAdnotata/OGA/tree/main/work_chronology}{https://github.com/OperaGraecaAdnotata/OGA\\/tree/main/work\_chronology}. 
It contains work and title metadata coming from the
canonical-greekLit\footnote{\href{https://github.com/PerseusDL/canonical-greekLit}{https://github.com/PerseusDL/canonical-greekLit}} and 
First1KGreek\footnote{\href{https://github.com/OpenGreekAndLatin/First1KGreek}{https://github.com/OpenGreekAndLatin/First1KGreek}} GitHub repositories 
and from the Perseus Catalogue.\footnote{\href{https://catalog.perseus.org/}{https://catalog.perseus.org/}}
The work dates, which follow the ISO 8601 format, 
were manually annotated by one annotator. 
All metadata should be considered work-in-progress.

\clearpage

\begin{table}
    \centering
    \begin{tabular}{|p{2.3cm}|p{2cm}|p{4cm}|>{\raggedleft\arraybackslash}p{3.2cm}|>{\raggedleft\arraybackslash}p{1.5cm}|}
    \hline
    \textbf{CTS} & \textbf{Author} & \textbf{Title} & \textbf{Date} & \textbf{Tokens} \\ \hline
        tlg0003.tlg001 & Thucydides & \makecell[l]{History of\\the Peloponnesian War} & \textminus0430-01/\textminus0410-12 & $32,344$ \\ \hline
        tlg0005.tlgxxx & Theocritus & Fragments & \textminus0299-01/\textminus0259-12 & $304$ \\ \hline
        tlg0006.tlg003 & Euripides & Medea & \textminus0430-01/\textminus0430-12 & $9,845$ \\ \hline
        tlg0007.tlg004 & Plutarch & Lycurgus & +0096-01/+0120-12 & $10,709$ \\ \cline{1-1} \cline{3-5}
        tlg0007.tlg015 &  & Alcibiades & +0096-01/+0120-12 & $11,439$ \\ \cline{1-1} \cline{3-5}
        tlg0007.tlg086 & & \makecell[l]{On the Fortunes\\of the Romans} & +0060-01/+0065-12 & $5,232$ \\ \cline{1-1} \cline{3-5}
        tlg0007.tlg087 & & \makecell[l]{On the Fortune or\\ the Virtue of\\Alexander I and II} & +0096-01/+0120-12 & $9,823$ \\ \hline
        tlg0008.tlg001 & \makecell[l]{Athenaeus\\of Naucratis} & The Deipnosophists & +0175-01/+0200-12 & $45,653$ \\ \hline
        tlg0009.tlg001 & Sappho & Fragments & \textminus0699-01/\textminus0599-12 & $4,530$ \\ \hline
        tlg0010.tlg002 & Isocrates & Against Callimachus & \textminus0401-01/\textminus0401-12 & 4,109 \\ \cline{1-1} \cline{3-5}
        tlg0010.tlg020 &  & To Philip & \textminus0345-01/\textminus0345-12 & $466$ \\ \hline
        tlg0011.tlg001 & Sophocles & Trachiniae & \textminus0449-01/\textminus0449-12 & $9,026$ \\ \cline{1-1} \cline{3-5}
        tlg0011.tlg002 & & Antigone & \textminus0442-01/\textminus0437-12 & $8,990$ \\ \cline{1-1} \cline{3-5}
        tlg0011.tlg003 &  & Ajax & \textminus0438-01/\textminus0435-12 & $9,751$ \\ \cline{1-1} \cline{3-5}
        tlg0011.tlg004 & & Oedipus Tyrannus & \textminus0418-01/\textminus0415-12 & $11,521$ \\ \cline{1-1} \cline{3-5}
        tlg0011.tlg005 & & Electra & \textminus0417-01/\textminus0406-12 & $10,806$ \\ \hline
        tlg0012.tlg001 & Homer & Iliad & \textminus0799-01/\textminus0700-12 & $130,479$ \\ \cline{1-1} \cline{3-5}
        tlg0012.tlg002 & & Odyssey & \textminus0799-01/\textminus0700-12 & $105,612$ \\ \hline
        tlg0013.tlg002 & Homeric Hymns & Hymn 2 to Demeter & \textminus0624-01/\textminus0574-12 & $3,968$ \\ \hline
        tlg0014.tlg001 & Demosthenes & First Olynthiac & \textminus0348-01/\textminus0348-12 & $2,194$ \\ \cline{1-1} \cline{3-5}
        tlg0014.tlg004 &  & First Philippic & \textminus0350-01/\textminus0350-12 & $3,951$ \\ \cline{1-1} \cline{3-5}
        tlg0014.tlg007 &  & On Halonnesus & \textminus0342-01/\textminus0341-12 & $2,886$ \\ \cline{1-1} \cline{3-5}
        tlg0014.tlg017 &  & \makecell[l]{On the Treaty\\with Alexander} & \textminus0330-01/\textminus0330-12 & $2,076$ \\ \cline{1-1} \cline{3-5}
        tlg0014.tlg018 &  & On the Crown & \textminus0329-01/\textminus0329-12 & $26,435$ \\ \cline{1-1} \cline{3-5}
        tlg0014.tlg027 &  & Against Aphobus I & \textminus0363-01/\textminus0362-12 & $5,346$ \\ \cline{1-1} \cline{3-5}
        tlg0014.tlg036 &  & For Phormio & \textminus0349-01/\textminus0348-12 & $4,649$ \\ \cline{1-1} \cline{3-5}
        tlg0014.tlg037 &  & Against Pantaenetus & \textminus0346-01/\textminus0346-12 & $4,528$ \\ \cline{1-1} \cline{3-5}
        tlg0014.tlg039 &  & Against Boeotus I & \textminus0347-01/\textminus0346-12 & $3,351$ \\ \cline{1-1} \cline{3-5}
        tlg0014.tlg041 &  & Against Spudias & \textminus0363-01/\textminus0358-12 & $2,333$ \\ \cline{1-1} \cline{3-5}
        tlg0014.tlg042 &  & Against Phaenippus & \textminus0329-01/\textminus0329-12 & $2,624$ \\ \cline{1-1} \cline{3-5}
        tlg0014.tlg045 &  & Against Stephanus I & \textminus0349-01/\textminus0348-12 & $6,839$ \\ \cline{1-1} \cline{3-5}
        tlg0014.tlg046 &  & Against Stephanus II & \textminus0349-01/\textminus0348-12 & $2,168$ \\ \cline{1-1} \cline{3-5}
        tlg0014.tlg047 &  & \makecell[l]{Against Evergus\\and Mnesibulus} & \textminus0354-01/\textminus0354-12 & $6,235$ \\ \cline{1-1} \cline{3-5}
        tlg0014.tlg049 &  & \makecell[l]{Apollodorus\\Against Timotheus} & \textminus0361-01/\textminus0361-12 & $5,005$ \\ \cline{1-1} \cline{3-5}
        tlg0014.tlg050 &  & \makecell[l]{Apollodorus\\Against Polycles} & \textminus0359-01/\textminus0359-12 & $5,306$ \\ \cline{1-1} \cline{3-5}
        tlg0014.tlg051 &  & \makecell[l]{On the Trierarchic Crown} & \textminus0359-01/\textminus0357-12 & $1,580$ \\ \cline{1-1} \cline{3-5}
        tlg0014.tlg052 &  & \makecell[l]{Apollodorus\\Against Callippus} & \textminus0368-01/\textminus0367-12 & $2,490$ \\ \cline{1-1} \cline{3-5}
        tlg0014.tlg053 &  & \makecell[l]{Apollodorus\\Against Nicostratus} & \textminus0367-01/\textminus0366-12 & $2,340$ \\ \hline       
    \end{tabular}
\end{table}

\clearpage

\begin{table}[!ht]
    \centering
    \begin{tabular}{|p{2.3cm}|p{2cm}|p{4cm}|>{\raggedleft\arraybackslash}p{3.2cm}|>{\raggedleft\arraybackslash}p{1.5cm}|}
    \hline
    \textbf{CTS} & \textbf{Author} & \textbf{Title} & \textbf{Date} & \textbf{Tokens} \\ \hline
        tlg0014.tlg054 & Demosthenes & Against Conon & \textminus0354-01/\textminus0340-12 & $3,755$ \\ \cline{1-1} \cline{3-5}
        tlg0014.tlg057 &  & Against Eubulides & \textminus0345-01/\textminus0344-12 & $5,498$ \\ \cline{1-1} \cline{3-5}
        tlg0014.tlg059 &  & \makecell[l]{Theomnestus and\\ Apollodorus\\Against Neaera} & \textminus0342-01/\textminus0339-12 & $10,489$ \\ \hline
        tlg0016.tlg001 & Herodotus & Histories & \textminus0429-01/\textminus0424-12 & $33,150$ \\ \hline
        tlg0017.tlg003 & Isaeus & The Estate of Pyrrhus & \textminus0388-01/\textminus0388-12 & $4,959$ \\ \hline
        tlg0019.tlg001 & Aristophanes & Acharnians & \textminus0424-01/\textminus0424-12 & $8,984$ \\ \cline{1-1} \cline{3-5}
        tlg0019.tlg008 &  & Thesmophoriazusae & \textminus0410-01/\textminus0410-12 & $9,073$ \\ \hline
        tlg0020.tlg001 & Hesiod & Theogony & \textminus0899-01/\textminus0700-12 & $8,234$ \\ \cline{1-1} \cline{3-5}
        tlg0020.tlg002 &  & Works and Days & \textminus0899-01/\textminus0700-12 & $7,116$ \\ \cline{1-1} \cline{3-5}
        tlg0020.tlg003 &  & Shield of Heracles & \textminus0899-01/\textminus0700-12 & $3,934$ \\ \hline
        tlg0026.tlg001 & Aeschines & Against Timarchus & \textminus0345-01/\textminus0344-12 & $15,971$ \\ \hline
        tlg0027.tlg001 & Andocides & On the Misteries & \textminus0399-01/\textminus0398-12 & $5,964$ \\ \hline
        tlg0028.tlg001 & Antiphon & \makecell[l]{Against the Stepmother\\for Poisoning} & \textminus0419-01/\textminus0410-12 & $2,046$ \\ \cline{1-1} \cline{3-5}
        tlg0028.tlg002 &  & First Tetralogy & \textminus0479-01/\textminus0410-12 & $2,915$ \\ \cline{1-1} \cline{3-5}
        tlg0028.tlg005 &  & On the Murder of Herodes & \textminus0417-01/\textminus0417-12 & $7,458$ \\ \cline{1-1} \cline{3-5}
        tlg0028.tlg006 &  & On the Choreutes & \textminus0418-01/\textminus0418-12 & $4,014$ \\ \hline
        tlg0032.tlg001 & Xenophon & Hellenica & \textminus0361-01/\textminus0353-12 & $27,401$ \\ \cline{1-1} \cline{3-5}
        tlg0032.tlg002 &  & Memorabilia & \textminus0409-01/\textminus0353-12 & $27,840$ \\ \cline{1-1} \cline{3-5}
        tlg0032.tlg004 &  & Symposium & \textminus0369-01/\textminus0360-12 & $7,291$ \\ \cline{1-1} \cline{3-5}
        tlg0032.tlg006 &  & Anabasis & \textminus0379-01/\textminus0359-12 & $18,737$ \\ \cline{1-1} \cline{3-5}
        tlg0032.tlg007 &  & Cyropaedia & \textminus0368-01/\textminus0365-12 & $50,690$ \\ \cline{1-1} \cline{3-5}
        tlg0032.tlg008 &  & Hiero & \textminus0356-01/\textminus0356-12 & $6,953$ \\ \cline{1-1} \cline{3-5}
        tlg0032.tlg015 &  & \makecell[l]{Constitution\\of the Athenians} & \textminus0442-01/\textminus0405-12 & $3,723$ \\ \hline
        tlg0041.tlg001 & Chion & Epistulae & +0001-01/+0100-12 & $5,577$ \\ \hline
        tlg0058.tlg001 & Aeneas Tacticus & Poliocetica & \textminus0374-01/\textminus0349-12 & $7,207$ \\ \hline
        tlg0059.tlg001 & Plato & Euthyphro & \textminus0398-01/\textminus0346-12 & $6,349$ \\ \cline{1-1} \cline{3-5}
        tlg0059.tlg002 &  & Apology & \textminus0398-01/\textminus0389-12 & $10,457$ \\ \cline{1-1} \cline{3-5}
        tlg0059.tlg003 &  & Crito & \textminus0398-01/\textminus0389-12 & $5,093$ \\ \cline{1-1} \cline{3-5}
        tlg0059.tlg029 &  & Cleiphon & \textminus0398-01/\textminus0346-12 & $1,875$ \\ \hline
        tlg0060.tlg001 & Diodorus of Sicily & Historical Library & \textminus0059-01/\textminus0029-12 & $25,692$ \\ \hline
        tlg0061.tlg001 & Lucian of Samosata & Asinus & +0125-01/+0180-12 & $11,054$ \\ \hline
        tlg0081.tlg001 & Dionysius of Halicarnassus & Antiquitates Romanae & \textminus0007-01/\textminus0006-12 & $30,312$ \\ \hline
        tlg0085.tlg001 & Aeschylus & Supplices & \textminus0465-01/\textminus0458-12 & $6,071$ \\ \cline{1-1} \cline{3-5}
        tlg0085.tlg002 &  & Persians & \textminus0471-01/\textminus0471-12 & $6,381$ \\ \cline{1-1} \cline{3-5}
        tlg0085.tlg003 &  & Prometheus Bound & \textminus0459-01/\textminus0455-12 & $7,222$ \\ \cline{1-1} \cline{3-5}
        tlg0085.tlg004 &  & Seven against Thebes & \textminus0466-01/\textminus0466-12 & $6,372$ \\ \cline{1-1} \cline{3-5}
        tlg0085.tlg005 &  & Agamemnon & \textminus0457-01/\textminus0457-12 & $10,037$ \\ \cline{1-1} \cline{3-5}
        tlg0085.tlg006 &  & Libation Bearers & \textminus0457-01/\textminus0457-12 & $5,846$ \\ \cline{1-1} \cline{3-5}
        tlg0085.tlg007 &  & Eumenides & \textminus0457-01/\textminus0457-12 & $6,518$ \\ \hline
    \end{tabular}
\end{table}

\clearpage

\begin{table}[!ht]
    \centering
    \begin{tabular}{|p{2.3cm}|p{2cm}|p{4cm}|>{\raggedleft\arraybackslash}p{3.2cm}|>{\raggedleft\arraybackslash}p{1.5cm}|}
    \hline
    \textbf{CTS} & \textbf{Author} & \textbf{Title} & \textbf{Date} & \textbf{Tokens} \\ \hline
        tlg0086.tlg035 & Aristotle & Politics & \textminus0399-01/\textminus0299-12 & $19,867$ \\ \hline
        tlg0093.tlg009 & Theophrastus & Characters & \textminus0316-01/\textminus0316-12 & $8,265$ \\ \hline
        tlg0096.tlg002 & Aesop & Aesop's Fables & \textminus0599-01/\textminus0500-12 & $5,221$ \\ \hline
        tlg0255.tlg001 & Mimnermus of Colophon & Fragmenta & \textminus0699-01/\textminus0599-12 & $213$ \\ \hline
        tlg0260.tlg001 & Semonides of Amorgos & Fragmenta & \textminus0699-01/\textminus0599-12 & $767$ \\ \hline
        tlg0343.tlg001 & Ezechiel & Exagogè & \textminus0199-01/\textminus0099-12 & $1,939$ \\ \hline
        tlg0429.tlg001 & Cephisodorus Comicus & Fragmenta & \textminus0401-01/\textminus0401-12 & $29,490$ \\ \hline
        tlg0526.tlg004 & Josephus Flavius & The Jewish War & +0075-01/+0075-12 & $24,987$ \\ \hline
        tlg0527.tlg001 & Septuaginta & Genesis & \textminus0299-01/\textminus0200-12 & $19,235$ \\ \hline
        tlg0537.tlg012 & Epicurus & Epistula ad Menoeceum & \textminus0310-01/\textminus0270-12 & $1,523$ \\ \hline
        tlg0540.tlg001 & Lysias & On the Murder of Eratosthenes & \textminus0402-01/\textminus0401-12 & $2,834$ \\ \cline{1-1} \cline{3-5}
        tlg0540.tlg012 &  & Against Eratosthenes & \textminus0402-01/\textminus0402-12 & $5,638$ \\ \cline{1-1} \cline{3-5}
        tlg0540.tlg013 &  & Against Agoratus & \textminus0399-01/\textminus0397-12 & $5,641$ \\ \cline{1-1} \cline{3-5}
        tlg0540.tlg014 &  & Against Alcibiades 1 & \textminus0394-01/\textminus0394-12 & $2,801$ \\ \cline{1-1} \cline{3-5}
        tlg0540.tlg015 & & Against Alcibiades 2 & \textminus0394-01/\textminus0394-12 & $688$ \\ \cline{1-1} \cline{3-5}
        tlg0540.tlg019 & & On the Property of Aristophanes & \textminus0386-01/\textminus0386-12 & $3,624$ \\ \cline{1-1} \cline{3-5}
        tlg0540.tlg023 & & Against Pancleon & \textminus0399-01/\textminus0398-12 & $896$ \\ \cline{1-1} \cline{3-5}
        tlg0540.tlg024 & & On the Refusal of a Pension & \textminus0402-01/\textminus0402-12 & $1,665$ \\ \hline
        tlg0541.tlg007 & Menander of Athens & Dyscolus & \textminus0315-01/\textminus0315-12 & $8,069$ \\ \hline
        tlg0543.tlg001 & Polybius & Histories & \textminus0167-01/\textminus0117-12 & $105,693$ \\ \hline
        tlg0544.tlg002 & Sextus Empiricus & Adversus Mathematicos & +0201-01/+0300-12 & $16,218$ \\ \hline
        tlg0548.tlg001 & Apollodorus & Library & +0101-01/+0200-12 & $1,265$ \\ \hline
        tlg0551.tlg017 & Appianus of Alexandria & Civil Wars & +0101-01/+0200-12 & $25,665$ \\ \hline
        tlg0554.tlg001 & Chariton & De Chaerea et Callirhoe & +0075-01/+0125-12 & $6,265$ \\ \hline
        tlg0557.tlg001 & Epictetus & Discourses & +0108-01/+0108-12 & $7,204$ \\ \hline
        tlg0559.tlg002 & Hero of Alexandria & De Automatis & +0062-01/+0085-12 & $10,321$ \\ \hline
        tlg0561.tlg001 & Longus & Daphnis and Chloe & +0101-01/+0300-12 & $672$ \\ \hline
        tlg0585.tlg001 & Phlegon of Tralles & Book of Marvels & +0100-01/+0200-12 & $5,642$ \\ \hline
        tlg1220.tlg001 & Batrachomyo-machia & Batrachomyomachia Homerica & \textminus0099-01/\textminus0029-12 & $2,212$ \\ \hline
        tlg2003.tlg001 & Julian & Panegyric in Honor of the Emperor Constantinus & +0355-01/+0355-12 & $1,405$ \\ \hline
        tlgxxxx.tlgxxx & Paeanius & Brevarium & +0337-01/+0379-12 & $6,184$ \\ \hline      
    \end{tabular}
\end{table}

\end{document}